\documentclass[10pt,twocolumn,letterpaper]{article}

\usepackage{amsmath}
\usepackage{tikz}
\usepackage{bm}
\usepackage{xcolor}

\usepackage{subcaption}
\usepackage{graphicx}
\usepackage{array}
\usepackage[export]{adjustbox}
\usepackage{algorithm}
\usepackage[noend]{algpseudocode}
\usepackage{tabularx}
\usepackage{booktabs}
\usepackage{makecell}

\usepackage[pagenumbers]{cvpr}

\definecolor{cvprblue}{rgb}{0.21,0.49,0.74}
\usepackage[pagebackref,breaklinks,colorlinks,allcolors=cvprblue]{hyperref}

\def\paperID{}
\def\confName{CVPR}
\def\confYear{2026}

\title{LightMedSeg: Lightweight 3D Medical Image Segmentation
       with Learned Spatial Anchors}

\author{Kavyansh Tyagi\\
National Institute of Technology Kurukshetra, India\\
{\tt\small 123109026@nitkkr.ac.in}
\and
Vishwas Rathi\thanks{$^*$Corresponding authors.}\\
National Institute of Technology Kurukshetra, India\\
{\tt\small vishwas@nitkkr.ac.in}
\and
Puneet Goyal$^*$\\
Indian Institute of Technology Ropar, India\\
{\tt\small puneet@iitrpr.ac.in}
}

\begin{document}
\maketitle
\begin{abstract}
Accurate and efficient 3D medical image segmentation is essential for clinical AI, where models must remain reliable under stringent memory, latency, and data availability constraints. Transformer-based methods achieve strong accuracy but suffer from excessive parameters, high FLOPs, and limited generalization. We propose LightMedSeg, a modular UNet-style segmentation architecture that integrates anatomical priors with adaptive context modeling. Anchor-conditioned FiLM modulation enables anatomy-aware feature calibration, while a local structural prior module and texture-aware routing dynamically allocate representational capacity to boundary-rich regions. Computational redundancy is minimized through ghost and depthwise convolutions, and multi-scale features are adaptively fused via a learned skip router with anchor-relative spatial position bias. Despite requiring only 0.48M parameters and 14.64~GFLOPs, LightMedSeg achieves segmentation accuracy within a few Dice points of heavy transformer baselines. Therefore, LightMedSeg is a deployable and data-efficient solution for 3D medical image segmentation. Code will be released  publicly upon acceptance.
\end{abstract}

\section{Introduction}
\label{sec:intro}

Deep learning has transformed medical image analysis, driving major gains in the accuracy and reliability of automated diagnostic systems. Within this domain, 3D medical image segmentation is foundational for tumor delineation, organ localization, and treatment planning. Conventional convolutional neural networks (CNNs), such as U-Net and its variants, provide strong baselines through hierarchical local feature extraction. However, their receptive field is inherently limited and restricts their ability to model global anatomical dependencies. This issue becomes critical in volumetric data with complex spatial structures and ambiguous boundaries.

To overcome these limitations, transformer based architectures have emerged, leveraging global self-attention to capture semantic coherence and contextual dependencies across entire volumes. Models such as nnFormer~\cite{nnFormer} and SegFormer3D~\cite{10678245} achieve state-of-the-art performance, often surpassing purely convolutional designs. Yet, these architectures are burdened by large parameter counts, high computational demands, and significant inference latency. Such constraints limit their practicality in real world clinical environments where computational resources are scarce and decisions must be made quickly.

Another persistent gap lies in the uniform way current methods process data. Most approaches overlook valuable anatomical priors, fail to adapt resources to regions of uncertainty or structural ambiguity, and rely on rigid skip connections with static fusion strategies. This undermines multi-scale integration, anatomical fidelity, and ultimately clinical reliability.

We propose LightMedSeg, a U-Net-based architecture designed for efficient 3D medical image segmentation under resource constraints. The model integrates several lightweight mechanisms that collectively achieve strong efficiency-accuracy trade-offs. Specifically, learned spatial anchors provide global context for feature conditioning via FiLM modulation, while an adaptive feature mixing module uses spatial gating to route features through multiple projection heads. In addition, a texture-aware routing mechanism separates high-variation regions from homogeneous areas to guide processing path selection, and a multi-scale skip fusion module learns to combine encoder features across scales rather than relying on fixed connections. Further, spatial position bias derived from anchor-relative distances provides explicit spatial guidance during decoding.

Despite its lightweight design, LightMedSeg achieves a competitive segmentation performance with only 0.48M parameters and 14.64 GFLOPs, making LightMedSeg significantly smaller than many traditional transformer-based models. By combining structural priors, adaptive feature routing, and lightweight global context modeling, LightMedSeg offers a practical solution for accurate and efficient volumetric segmentation, narrowing the gap between research prototypes and deployable clinical systems. Our contributions are summarized as follows: (i) we introduce LightMedSeg, a lightweight 3D medical image segmentation architecture having only 0.48M parameters and maintaining competitive accuracy, (ii) we propose a module that utilizes specific spatial anchors and employs global context using anchor conditioned FiLM modulation, (iii) we design a local structural prior module (LSPM) that identifies structurally complex regions and guides features through appropriate processing paths, (iv) we introduce a learned multi scale skip fusion design that adaptively combines encoder features from different scales instead of using fixed U-Net skip connections, and (v) experiments on BraTS and ACDC show that LightMedSeg achieves a strong balance between segmentation accuracy and computational efficiency.
\section{Related Work}
\label{sec:rel}

Recent progress in 3D medical image segmentation has explored convolutional networks, transformers, and hybrid architectures, each aiming to balance accuracy, efficiency, and generalization. CNNs remain effective for local feature capture, transformers excel at global context modeling, and hybrid designs attempt to unify both.

The U-Net~\cite{21/7785132} and its descendants, including deeply supervised CNNs~\cite{33/7965852}, DenseUNet~\cite{2/QIMS43519}, and volumetric variants such as 3D U-Net~\cite{6/çiçek20163dunetlearningdense}, V-Net~\cite{21/7785132}, and nnU-Net~\cite{13/isensee2018nnunetselfadaptingframeworkunetbased}, established the foundation for volumetric segmentation through hierarchical multi-scale features. However, their inherently local receptive fields limit modeling of long-range anatomical dependencies. Extensions via dilated convolutions, pyramid pooling, and attention gates~\cite{chen2024frequency, wu2022p2t, chen2021end} improved context modeling, but still remained bounded by convolutional locality and scaled poorly with deeper backbones~\cite{2a}.

Transformer-based designs introduced global self-attention to address these shortcomings. SETR~\cite{zheng2021rethinkingSETR} pioneered a sequence-to-sequence formulation but suffered from spatial detail loss during patch embedding. Hybrid designs such as TransBTS~\cite{TransBTS}, CoTr~\cite{DBLP:journals/corr/abs-2012-15840}, TransFuse~\cite{zhang2021transfuse}, and LoGoNet~\cite{logonetkarimi2024masked} integrated CNN encoders with transformer modules using cross-attention or deformable attention~\cite{29/xia2022vision}. While improving local global fusion, these pipelines were computationally heavy and often dataset specific.

Hierarchical transformer-only networks like Swin-UNet~\cite{swinUNet} and SwinUNETR~\cite{10.1007/978-3-031-08999-2_22} captured multi-resolution features via shifted window attention, but window partitioning restricted long-range anatomical modeling. Efficient variants such as LeViT-UNet~\cite{LeViT-UNet} and SegFormer3D~\cite{10678245} reduced parameters with lightweight decoders and streamlined attention, with SegFormer3D reaching 4.51M parameters and 17.5 GFLOPs. Yet these still relied on static skip fusion and struggled under anatomical variation or low contrast conditions.

State-of-the-art designs such as nnFormer~\cite{nnFormer}, which replaced concatenation with skip attention, and UNETR~\cite{UNETR}, which employed full transformer encoders, significantly improved context capture but at high computational cost. UNETR++~\cite{unetr++} reduced overhead through multi-scale cross skip connections and lightweight decoders, but remained parameter-heavy and reliant on dense supervision.

Despite these advances, existing approaches remain challenged by static feature fusion, large parameterization, and dependence on voxel-level supervision.

In summary, several methods have reduced computational overhead and improved accuracy, they remain limited by rigid fusion, heavy parameter counts, and sensitivity to annotation scarcity. LightMedSeg addresses all three challenges simultaneously, offering anatomically-aware segmentation that is efficient, adaptive, and reliable under real-world constraints.

\section{Proposed Method}
\label{sec:method}

\begin{figure*}[!t]
  \centering
  \includegraphics[width=0.9\textwidth]{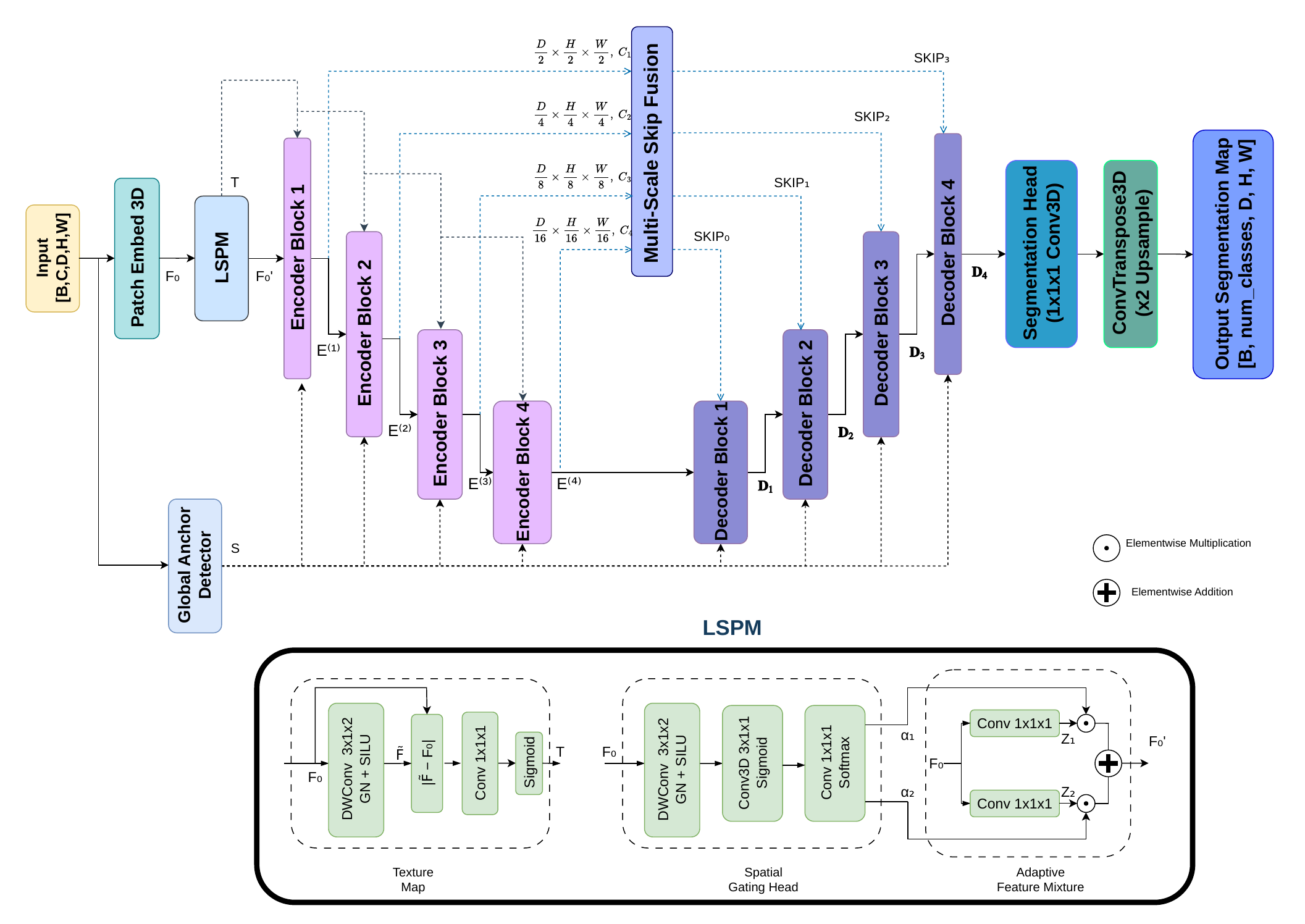}
\caption{%
Architecture of \textbf{LightMedSeg}. The input volume is
first embedded by a stride-2 GhostConv3D stem. The Global Anchor
Detector and the Local Structural Prior Module (LSPM) extract
spatial anchors~$\mathbf{S}$ and a texture routing
map~$\mathbf{T}$ from the stem features. These priors guide
(i)~adaptive feature mixing before the encoder,
(ii)~anchor-conditioned, texture-routed processing at every encoder
stage, (iii)~multi-scale learned skip fusion, and (iv)~an adaptive
decoder with content-tied spatial position bias. A final
ConvTranspose3D restores voxel-wise class logits at the original
resolution.}

  \label{fig:lightmedseg}
\end{figure*}

LightMedSeg is designed for 3D medical image segmentation in resource constraints environment, where memory and compute budgets preclude large transformer-based models. Given an input volume
$\mathbf{X}\in\mathbb{R}^{B\times C_{\mathrm{in}}\times D\times H\times W}$,
the network produces voxel-wise class logits
$\hat{\mathbf{Y}}\in\mathbb{R}^{B\times N_{\mathrm{cls}}\times
D\times H\times W}$.

LightMedSeg follows a U-Net style encoder-decoder design, where
computational redundancy is minimized through ghost and depthwise
convolutions while introducing structural context through lightweight prior modules.  A learned skip router and
anchor-relative spatial position bias in the decoder recover
fine-grained anatomical detail without increasing model complexity.

The overall architecture comprises five components:(i)~a GhostConv3D patch embedding stem that halves spatial resolution and produces modality independent feature maps, (ii)~a global anchor detector that predicts sample specific spatial anchors from the input, (iii)~a local structural prior module~(LSPM) that extracts a texture routing map and a structurally adapted feature map, (iv)~a four-stage encoder with anchor-conditioned FiLM modulation and texture-aware routing at each stage, and (v)~a symmetric four-stage decoder with learned multi-scale skip fusion, factored spatial position bias, and adaptive multi path processing.  Each component is described in detail below.

\subsection{Patch Embedding Stem}
\label{sec:stem}

The input volume $\mathbf{X} \in \mathbb{R}^{B \times C_{\mathrm{in}}
\times D \times H \times W}$ has $C_{\mathrm{in}}$ channels,
corresponding to the number of imaging modalities. Standard 3D patch embedding
projects this directly through a dense Conv3D, incurring
higher parameters regardless of spatial redundancy. We instead adopt GhostConv3D~\cite{han2020ghostnet}
with ghost ratio $\rho\!=\!2$, which splits feature generation into
two steps: a strided Conv3D with kernel size 3 and stride 2 produces
half the required output channels from the input as primary features,
and a depthwise Conv3D with kernel size 3 and stride 1 synthesizes
the remaining half from those primary features as ghost features.
The two are concatenated and normalized via Group Normalization with
four groups ($\mathrm{GN}_4$), yielding the stem output
$\mathbf{F}_0 \in \mathbb{R}^{B \times C_0 \times \frac{D}{2}
\times \frac{H}{2} \times \frac{W}{2}}$ with $C_0\!=\!8$,
approximately $\!2\times$ fewer parameters and FLOPs than a standard
Conv3D stem.

The stride 2 reduction ($128^3$ to $64^3$ for BraTS inputs) reduces
the memory footprint of all downstream operations and is fully
recovered by the four transposed convolution stages in the decoder
(Section~\ref{sec:decoder}). The output is a fixed 8 channel feature
volume irrespective of the number of input modalities, providing a
consistent representation to all downstream modules.

\subsection{Global Anchor Detector}
\label{sec:anchors}

Global context in purely convolutional encoders relies on large
receptive fields, which are prohibitively expensive at 3D scale,
while attention-based alternatives incur quadratic voxel complexity.
We instead propose a lightweight \emph{Global Anchor Detector} that
distils global spatial information from the raw input $\mathbf{X}$
into $K\!=\!8$ anchor coordinates, each normalized to $[0,1]^3$
along the depth, height, and width axes, representing salient
spatial locations in the volume. Anchors are predicted without
any spatial supervision at a cost of only 8.6K parameters.

The detector uses a three-layer encoder $\phi$ (Conv3D, GN$_4$,
SiLU; kernel 3, stride 2, padding 1) that progressively compresses
$\mathbf{X}$ into a compact feature volume. A 3D Global Average
Pooling layer then collapses spatial dimensions into a per-sample
descriptor, passed to a two-layer MLP (hidden width 128, SiLU)
that projects to $3K$ logits. A sigmoid activation constrains
each coordinate to $[0,1]$, ensuring all anchors lie within the
input volume:
\begin{equation}
  \mathbf{S} =
      \sigma\!\left(\mathrm{MLP}\!\left(
      \mathrm{GAP3D}(\phi(\mathbf{X}))\right)\right)
      \in [0,1]^{B\times K\times 3},
  \label{eq:anchors}
\end{equation}
where each of the $K$ rows of $\mathbf{S}$ encodes the normalized coordinate $(d,h,w)$ of one predicted anchor. Anchors are
sample specific as they adapt to each individual input volume rather
than being fixed positional tokens. $\mathbf{S}$ serves two roles
downstream: FiLM-based feature modulation at every encoder stage
(Section~\ref{sec:encoder}), and content-adaptive spatial position
bias throughout the decoder (Section~\ref{sec:decoder}).

\subsection{Local Structural Prior Module (LSPM)}
\label{sec:lspm}

Standard convolutional encoders apply identical computation to
every voxel, wasting capacity on homogeneous interiors while
potentially under-serving boundaries. The Local Structural Prior
Module (LSPM) addresses this via a lightweight three-branch block
operating on the stem output $\mathbf{F}_0$, emitting two signals:
a per-voxel texture routing map $\mathbf{T}$ consumed by all four
encoder stages, and a structurally adapted feature map
$\mathbf{F}_0'$ passed as the encoder's initial input, at a cost
of only 11.7K parameters (2.4\% of the model).

The three branches operate in parallel: Branch~1 detects
\emph{where} structural complexity exists via high-frequency
intensity transitions; Branch~2 estimates \emph{how much}
complexity is present in the learned feature space; and Branch~3
consumes this estimate to adaptively blend two parallel feature
projections, favouring expressive capacity near boundaries and
simpler processing in smooth interiors.

Texture Map (Branch 1):
Boundary and interface voxels in medical volumes exhibit sharp local
intensity transitions as tissue interiors are largely smooth. To
identify these structurally rich regions without a feature pyramid,
a $5\!\times\!5\!\times\!5$ depthwise convolution where $s{=}1$, $p{=}2$,
$\mathrm{groups}{=}C_0$, smooths $\mathbf{F}_0$. So the absolute
residual isolates high frequency content:
\begin{align}
  \tilde{\mathbf{F}}
    &= \mathrm{SiLU}\!\left(\mathrm{GN}_4\!\left(
       \mathrm{DWConv3D}\!\left(\mathbf{F}_0;\right.\right.\right.
       \nonumber \\
    &\quad \left.\left.\left.
       k{=}5,\;s{=}1,\;p{=}2,\;\mathrm{groups}{=}C_0
       \right)\right)\right),
    \label{eq:dwconv_tex}\\[4pt]
  \mathbf{T}
    &= \sigma\!\left(
       \mathrm{Conv}_{1\times1\times1}\!\left(
       \left|\tilde{\mathbf{F}}-\mathbf{F}_0\right|;\;
       C_0\!\to\!1\right)\right) \nonumber \\
    &\quad \in [0,1]^{B\times1\times64^3}.
    \label{eq:tex_map}
\end{align}
$\mathbf{T}_{b,d,h,w}$ at boundaries and interfaces. $\mathbf{T}$ is forwarded
unchanged to all four encoder stages as a stable routing reference.
This branch costs $1$K parameters.

Spatial Gating Head (Branch 2):
While $\mathbf{T}$ captures high-frequency texture in the input
space, it does not directly reflect structural complexity in the
learned feature space. We therefore estimate a complementary
per-voxel complexity score $\mathbf{G}$ via a two-layer
convolutional head $\psi$. Two consecutive convolutions with kernel size 3, stride 1 and padding 1, each followed by $\mathrm{GN}_4$ and SiLU, progressively
expand $\mathbf{F}_0$ from $C_0$ channels to $2C_0$ channels and
then collapse to a single-channel output, giving $\psi$ an effective
receptive field of $5\!\times\!5\!\times\!5$ voxels:

\begin{equation}
  \mathbf{G} = \sigma(\psi(\mathbf{F}_0))
      \in [0,1]^{B\times1\times64^3}.
  \label{eq:gating_map}
\end{equation}
Values near~1 indicate structurally complex voxels; values near~0
indicate smooth regions. $\mathbf{G}$ is passed internally to
Branch~3, where it is projected to per-voxel expert weights
$\boldsymbol{\alpha}_1, \boldsymbol{\alpha}_2$ via a
$1\!\times\!1\!\times\!1$ convolution followed by softmax, satisfying
$\alpha_{1,\mathbf{r}}+\alpha_{2,\mathbf{r}}=1$ at every voxel
$\mathbf{r}$. This branch costs $10.5$K parameters.

Adaptive Feature Mixer (Branch 3):
The stem applies a fixed linear projection uniformly to all voxels,
regardless of local complexity. We address this with a soft mixture
of two parallel $1\!\times\!1\!\times\!1$ expert projections
$\mathbf{Z}_1$ and $\mathbf{Z}_2$, each mapping from $C_0$ to $C_0$
channels, recombined in spatially varying proportions:
\begin{equation}
  \mathbf{F}_0' =
      \boldsymbol{\alpha}_1\odot\mathbf{Z}_1
      + \boldsymbol{\alpha}_2\odot\mathbf{Z}_2
      \in \mathbb{R}^{B\times C_0\times64^3}.
  \label{eq:mix}
\end{equation}
The softmax constraint ensures the full representational budget is
redistributed between experts rather than reduced. The
$1\!\times\!1\!\times\!1$ kernels restrict each expert to channel
recombination only, leaving all spatial processing to the encoder.
Expert specialization between complex and smooth regions emerges
from end-to-end training without explicit supervision, at a cost of
only $0.15$K parameters.

The LSPM takes
$\mathbf{F}_0\in\mathbb{R}^{B\times C_0\times64^3}$ as its sole
input and emits $\mathbf{T}\in[0,1]^{B\times1\times64^3}$ and
$\mathbf{F}_0'\in\mathbb{R}^{B\times C_0\times64^3}$ which are fed
into the encoder stages.

\subsection{Encoder Hierarchy}
\label{sec:encoder}

The encoder is a four-stage hierarchical feature extractor
progressively abstracting the input from fine-grained local
structure to coarse semantics. At each stage $i\!\in\!\{1,2,3,4\}$,
three operations are applied: anchor-conditioned $\mathrm{GhostConv3D}$
projection via FiLM~\cite{perez2018film}, which scales and shifts
feature channels by the learned anchor coordinates; texture-aware
dual-branch routing; and SE channel recalibration. Each stage output
is stored as skip feature $\mathbf{E}^{(i)}$ before downsampling.

Stages~1 to 3 apply MaxPool3D with kernel size 2 and stride 2 after
feature storage to halve all spatial dimensions. The stage~4 retains the
bottleneck resolution via an identity mapping, as further downsampling would destroy the spatial detail required by the decoder during reconstruction.

The encoder receives $\mathbf{F}_0'\in\mathbb{R}^{B\times
C_0\times\frac{D}{2}\times\frac{H}{2}\times\frac{W}{2}}$ from the
LSPM as its stage 1 input. Channel widths follow the schedule
$C_i\!\in\!\{8,16,32,64\}$, doubling at each stage, while spatial
resolutions progressively reduce from $\frac{D}{2}$ down to
$\frac{D}{4}$, $\frac{D}{8}$, and finally $\frac{D}{16}$ along the
depth axis (similarly for $H$ and $W$). The bottleneck representation
$\mathbf{E}^{(4)}\!\in\!\mathbb{R}^{B\times64\times
\frac{D}{16}\times\frac{H}{16}\times\frac{W}{16}}$ seeds the decoder.

Purely local convolutional encoders have no mechanism to incorporate
global spatial context, which is critical for localizing
anatomically specific structures in medical volumes. 
We therefore condition every encoder stage on the learned spatial
anchors via FiLM modulation. The anchor coordinates $\mathbf{S}$ are
first flattened into a single vector $\mathbf{s}$, which is then
passed through two independent linear projections
$\mathbf{W}_{\gamma,i}\!\in\!\mathbb{R}^{C_i\times(K\times3)}$ and
$\mathbf{W}_{\beta,i}\!\in\!\mathbb{R}^{C_i\times(K\times3)}$,
whose weights are learned during training, to produce a per-stage
scale parameter $\boldsymbol{\gamma}_i = \mathbf{W}_{\gamma,i}\mathbf{s}$
and shift parameter $\boldsymbol{\beta}_i = \mathbf{W}_{\beta,i}\mathbf{s}$,
each of dimension $C_i$, providing one scale and one shift value per
feature channel. These modulate the GhostConv3D output at stage $i$ to produce the
conditioned feature map $\mathbf{F}_i^{(\mathrm{cond})}
\in\mathbb{R}^{B\times C_i\times D_{i-1}\times H_{i-1}\times W_{i-1}}$:
\begin{equation}
  \mathbf{F}_i^{(\mathrm{cond})}
    = \bigl(1 + \boldsymbol{\gamma}_i\bigr)
       \circledast \mathrm{GhostConv3D}\!\left(\mathbf{F}_{i-1}\right)
       + \boldsymbol{\beta}_i,
  \label{eq:film_apply}
\end{equation}

The conditioned features $\mathbf{F}_i^{(\mathrm{cond})}$ are then
processed through a texture-aware two-path routing step, where
the routing is governed by $\mathbf{T}_i$, generated by texture map
$\mathbf{T}$ trilinearly resampled to the current stage resolution
$(D_{i-1},H_{i-1},W_{i-1})$. Rather than applying a single
convolution uniformly across all voxels, the two paths
specialize: the detail-preserving path produces
$\mathbf{Z}_i^{\mathrm{detail}}\in\mathbb{R}^{B\times C_i\times
D_{i-1}\times H_{i-1}\times W_{i-1}}$ by capturing fine-grained
boundary structure via a depthwise convolution over a local spatial
neighborhood, while the smoothing path produces
$\mathbf{Z}_i^{\mathrm{smooth}}\in\mathbb{R}^{B\times C_i\times
D_{i-1}\times H_{i-1}\times W_{i-1}}$ by handling homogeneous
interior regions cheaply via a $1\!\times\!1\!\times\!1$ pointwise
convolution. Formally:

\begin{align}
  \mathbf{Z}_i^{\mathrm{detail}}
    &= \mathrm{SiLU}\!\left(\mathrm{GN}_4\!\left(
       \mathrm{DWConv}_{3^3}\!\left(
       \mathbf{F}_i^{(\mathrm{cond})};\right.\right.\right.
       \nonumber\\
    &\quad\left.\left.\left.
       s{=}1,\;p{=}1,\;\mathrm{groups}{=}C_i
       \right)\right)\right),
    \label{eq:detail_branch}\\[4pt]
  \mathbf{Z}_i^{\mathrm{smooth}}
    &= \mathrm{Conv}_{1\times1\times1}\!\left(
       \mathbf{F}_i^{(\mathrm{cond})};\;
       C_i\!\to\!C_i,\;s{=}1\right),
    \label{eq:smooth_branch}
\end{align}
and blend them continuously using $\mathbf{T}_i$ as a soft gate:
\begin{equation}
  \mathbf{F}_i^{(\mathrm{tex})} =
      \mathbf{T}_i \circledast \mathbf{Z}_i^{\mathrm{detail}}
      + (1-\mathbf{T}_i) \circledast \mathbf{Z}_i^{\mathrm{smooth}},
  \label{eq:tex_blend}
\end{equation}
where $\mathbf{T}_i\!\in\![0,1]^{B\times1\times D_{i-1}\times
H_{i-1}\times W_{i-1}}$ is broadcast over $C_i$ channels.

Finally, after the two-path routing, a Squeeze-and-Excitation
block~\cite{hu2018senet} recalibrates the per-channel responses
of $\mathbf{F}_i^{(\mathrm{tex})}$ via global average pooling and
a bottleneck MLP:
\begin{equation}
  \begin{split}
  \mathbf{E}^{(i)}
    &= \sigma\!\left(\mathbf{W}_{2,i}\cdot\mathrm{SiLU}\!\left(
       \mathbf{W}_{1,i}\cdot\mathrm{GAP3D}\!\left(
       \mathbf{F}_i^{(\mathrm{tex})}\right)\right)\right)\\
    &\quad \circledast\,\mathbf{F}_i^{(\mathrm{tex})},
  \end{split}
  \label{eq:se_refine}
\end{equation}
where $\mathbf{W}_{1,i}\!\in\!\mathbb{R}^{b_i\times C_i}$,
$\mathbf{W}_{2,i}\!\in\!\mathbb{R}^{C_i\times b_i}$, and
$b_i\!=\!\max(4,\lfloor C_i/8\rfloor)$. The minimum bottleneck
width of~4 prevents channel-agnostic collapse at the narrow early
stages. $\mathbf{E}^{(i)}$ is stored for skip fusion and passed
through MaxPool3D to seed stage $i\!+\!1$.
\subsection{Multi-Scale Skip Fusion}
\label{sec:skip_router}

Standard U-Net skip connections forward only same-scale encoder
features, discarding all cross-scale context. Inspired by
full-scale skip designs~\cite{huang2020unet3plus}, we replace
this with a learned router that adaptively combines all four
encoder stages at each decoder level via per-voxel softmax
weights, at a cost of only 24K parameters.

Since encoder stages produce features of different widths
($C_i\!\in\!\{8,16,32,64\}$), each $\mathbf{E}^{(i)}$ is first
projected to a common width of 64 via a $1\!\times\!1\!\times\!1$
convolution before routing for $i\!\in\!\{1,2,3,4\}$:
\begin{equation}
  \hat{\mathbf{E}}^{(i)}
    = \mathrm{Conv}_{1\times1\times1}\!\left(\mathbf{E}^{(i)}\right)
    \in\mathbb{R}^{B\times64\times D_{i-1}\times H_{i-1}\times W_{i-1}},
  \label{eq:align_proj}
\end{equation}

At the two higher-resolution decoder stages ($j\!=\!3,4$),
concatenating all four aligned maps is expensive
in memory. Only the spatially corresponding encoder feature map
$\mathbf{E}^{(S-1-j)}$ is therefore projected to $C_j$ channels
and trilinearly upsampled to $(D_j,H_j,W_j)$, where the index
$(M\!-\!1\!-\!j)$ selects encoder stage~1 for $j\!=\!3$ and the
stem output for $j\!=\!4$, and this single aligned map is used
directly as $\mathbf{SKIP}_j$ at those stages.

At the two lower-resolution stages where $j\!=\!1$ and $j\!=\!2$,
the full learned router is activated. Each $\hat{\mathbf{E}}^{(i)}$
is trilinearly upsampled to $(D_j, H_j, W_j)$ and all four are
concatenated into a tensor of 256 channels. A two-layer CNN
controller consisting of two $1\!\times\!1\!\times\!1$ convolutions
with a SiLU activation in between, first reducing from 256 to 64
channels and then projecting to $M$ channels, predicts per-voxel
logits $\ell_{j,i,\mathbf{r}}$ over encoder stages. Softmax
normalization gives routing weights and the fused signal:

\begin{align}
  \boldsymbol{\alpha}_{j,i,\mathbf{r}}
    &= \frac{e^{\ell_{j,i,\mathbf{r}}}}
            {\displaystyle\sum_{i'=1}^{M}
             e^{\ell_{j,i',\mathbf{r}}}},
       \quad
       \sum_{i=1}^{M}\boldsymbol{\alpha}_{j,i,\mathbf{r}}\!=\!1,
    \label{eq:routing_weights}\\[4pt]
  \mathbf{SKIP}_j
    &= \sum_{i=1}^{M}\boldsymbol{\alpha}_{j,i}
       \odot\hat{\mathbf{E}}^{(i)}_j \nonumber\\
    &\quad\in\mathbb{R}^{B\times64\times
       D_j\times H_j\times W_j},
    \label{eq:skip_fusion}
\end{align}

where $\boldsymbol{\alpha}_{j,i}\!\in\!\mathbb{R}^{B\times1\times
D_j\times H_j\times W_j}$ is broadcast over 64 channels. A final $1\!\times\!1\!\times\!1$ convolution projects the fused features from 64 channels down to $C_j$ channels, yielding
$\mathbf{SKIP}_j \in \mathbb{R}^{B\times C_j\times D_j\times H_j\times W_j}$,
which is consumed by decoder stage $j$.

\subsection{Adaptive Decoder}
\label{sec:decoder}

Standard convolutional decoders apply uniform spatial operations
across all positions and lack awareness of anatomically significant
structures. We address both limitations through a four-stage decoder
combining anchor-relative spatial position bias, multi-scale skip
fusion, and per-voxel adaptive path selection.

The decoder is seeded by a $1\!\times\!1\!\times\!1$ bottleneck
projection of $\mathbf{E}^{(4)}$, producing
$\mathbf{F}_{\mathrm{dec}}\!\in\!\mathbb{R}^{B\times64\times
\frac{D}{16}\times\frac{H}{16}\times\frac{W}{16}}$ as input to
the first stage. At each stage $j\!\in\!\{1,2,3,4\}$, three
operations are applied sequentially by position-biased upsampling
and skip fusion, adaptive multi-path processing, and SE channel
recalibration. Channel widths follow a mirrored schedule
$C_j\!\in\!\{64,32,16,8\}$, halving as resolution doubles. The
stage input is upsampled by a $2\!\times\!2\!\times\!2$ transposed
convolution, yielding
$\mathbf{U}_j\!\in\!\mathbb{R}^{B\times C_j\times D_j\times
H_j\times W_j}$.

Since transposed convolutions carry no anatomical awareness, we
add a Spatial Position Bias $\mathbf{SPB}_j\!\in\!\mathbb{R}^{
B\times C_j\times D_j\times H_j\times W_j}$ to $\mathbf{U}_j$
before skip fusion. Naively storing the full 3D displacement
$(\mathbf{r}-\mathbf{s}_k)$, where $\mathbf{r}\!=\!(d,h,w)\!
\in\!\mathbb{Z}^3$ is a voxel coordinate and $\mathbf{s}_k\!\in\!
[0,1]^3$ is the normalized anchor location. This would require a
$B\!\times\!3K\!\times\!D_j\!\times\!H_j\!\times\!W_j$ tensor.
We instead decompose it axis-by-axis and reconstruct via
broadcasting, reducing peak memory by $3$ times.

We construct a per voxel signed distance map
$\mathbf{P}_{j,k}\!\in\!\mathbb{R}^{D_j\times H_j\times W_j}$,
where each entry $\mathbf{P}_{j,k}[d,h,w]$ stores the cumulative
signed offset of voxel $(d,h,w)$ from anchor $k$ along all three
spatial axes. For a voxel at integer index
$(d,h,w)$ with $d\!\in\!\{0,\ldots,D_j\!-\!1\}$,
$h\!\in\!\{0,\ldots,H_j\!-\!1\}$, $w\!\in\!\{0,\ldots,W_j\!-\!1\}$,
it is given as:
\begin{equation}
  \mathbf{P}_{j,k}[d,h,w] =
    \left(\frac{d}{D_j}-s_{k,d}\right)
    +\left(\frac{h}{H_j}-s_{k,h}\right)
    +\left(\frac{w}{W_j}-s_{k,w}\right)
  \label{eq:factored_pos}
\end{equation}
which decomposes the full 3D displacement into three independent
1D arrays having one per axis and are summed elementwise.

Stacking over all $K$ anchors gives a $B\!\times\!K\!\times\!
D_j\!\times\!H_j\!\times\!W_j$ volume, which is projected to
$C_j$ channels via a $1\!\times\!1\!\times\!1$ convolution to
produce $\mathbf{SPB}_j$ (Eq.~\ref{eq:spb}). Unlike fixed
sinusoidal or learnable grid encodings, $\mathbf{SPB}_j$ changes
with every input volume, giving the decoder dynamic positional
awareness tied to the predicted anchor locations $\mathbf{S}$.

\begin{align}
  \mathbf{SPB}_j
    &= \mathrm{Conv}_{1\times1\times1}\!\left(
       \mathrm{stack}\!\left(
       \{\mathbf{P}_{j,k}\}_{k=1}^{K},\,\mathrm{dim}{=}1
       \right);\right. \nonumber\\
    &\quad\left.
       K\!\to\!C_j,\;s{=}1
       \right)
       \in\mathbb{R}^{B\times C_j\times
       D_j\times H_j\times W_j},
    \label{eq:spb}
\end{align}

The position biased upsampled features and the encoder skip signal
$\mathbf{SKIP}_j$ are concatenated and passed through a $\mathrm{Conv}_{1\times1\times1}$ to produce the fused representation:
\begin{equation}
  \mathbf{D}_j^{(\mathrm{in})} =
      \mathrm{Conv}_{1\times1\times1}\!(
      \mathrm{Concat}\![
      \mathbf{U}_j\!+\!\mathbf{SPB}_j,\;\mathbf{SKIP}_j])
  \label{eq:decoder_fusion}
\end{equation}
to produce fused representation $\mathbf{D}_j^{(\mathrm{in})}$, which recalibrates the
fused channels before multi-path processing.

The fused representation is processed by three complementary parallel branches, each applying a different spatial transformation. A shared $1\times1\times1$ convolution produces
per-voxel soft blending weights $\boldsymbol{\pi}_{j,p}$ via softmax:
\begin{equation}
  \boldsymbol{\pi}_{j,p} = \mathrm{softmax}_p\!\left(
    \mathrm{Conv}_{1\times1\times1}\!\left(
    \mathbf{D}_j^{(\mathrm{in})};\; C_j \to P\right)\right),
\end{equation}
where $\sum_{p=1}^{P} \pi_{j,p,\mathbf{r}} = 1$ at every voxel
$\mathbf{r}$. The final gated output is their weighted sum:
\begin{equation}
  \mathbf{D}_j^{(\mathrm{gated})} = \sum_{p=1}^{P}\,
    \boldsymbol{\pi}_{j,p} \odot \mathcal{F}_p\!\left(
    \mathbf{D}_j^{(\mathrm{in})}\right),
\end{equation}
where each $\mathcal{F}_p$ denotes one of the three branches:
\begin{align}
  \mathcal{F}_1\!\left(\mathbf{D}_j^{(\mathrm{in})}\right)
    &= \mathrm{SiLU}\!\left(\mathrm{GN}_4\!\left(
       \mathrm{DWConv3D}_{3}\!\left(
       \mathbf{D}_j^{(\mathrm{in})}\right)\right)\right), \\[4pt]
  \mathcal{F}_2\!\left(\mathbf{D}_j^{(\mathrm{in})}\right)
    &= \mathrm{SiLU}\!\left(\mathrm{GN}_4\!\left(
       \mathrm{GhostConv3D}_{3,\,\rho=2}\!\left(
       \mathbf{D}_j^{(\mathrm{in})}\right)\right)\right), \\[4pt]
  \mathcal{F}_3\!\left(\mathbf{D}_j^{(\mathrm{in})}\right)
    &= \mathrm{SiLU}\!\left(\mathrm{GN}_4\!\left(
       \mathrm{Conv}_{1\times1\times1}\!\left(
       \mathbf{D}_j^{(\mathrm{in})}\right)\right)\right).
\end{align}

$\mathcal{F}_1$ (depthwise Conv3D, kernel $3$, groups $C_j$)
aggregates local neighbourhoods, making it effective at organ
boundaries; $\mathcal{F}_2$ (GhostConv3D, kernel $3$,
$\rho\!=\!2$) captures multi-scale context at half the cost of
a standard Conv3D; and $\mathcal{F}_3$ ($1\!\times\!1\!\times\!1$
conv) performs pure channel mixing suited to smooth homogeneous
regions. The blending weights allow each voxel to adaptively
select its own mixture of local, multi-scale, and channel-only
processing.

The gated output $\mathbf{D}_j^{(\mathrm{gated})}$ is then passed
through a Squeeze-and-Excitation block with bottleneck width
$b_j = \max\!\left(4,\lfloor C_j/8 \rfloor\right)$, which
recalibrates channel importance to produce the final stage output
$\mathbf{D}_j \in \mathbb{R}^{B \times C_j \times D_j \times H_j \times W_j}$.

The four stages progressively restore
resolution from $\frac{D}{16}$ through $\frac{D}{8}$, $\frac{D}{4}$,
and $\frac{D}{2}$ back to the full input resolution $D$, with the
final output $\mathbf{D}_4\!\in\!\mathbb{R}^{B\times C_0\times
D\times H\times W}$ carrying $C_0$ (here $C_0\!=\!8$) channels.

A $1\!\times\!1\!\times\!1$ convolution first maps $\mathbf{D}_4$
to $N_{\mathrm{cls}}$ channels, followed by a $2\!\times\!2\!\times\!2$
transposed convolution with stride~2 that restores the full input
resolution, yielding voxel-wise class logits
$\hat{\mathbf{Y}}\!\in\!\mathbb{R}^{B\times N_{\mathrm{cls}}
\times D\times H\times W}$, passed to the loss at training time
and to $\arg\max$ at inference.

\subsection{Training Objective}
\label{sec:loss}

We train with a weighted combination of Dice, cross-entropy, and
boundary losses:
\begin{equation}
  \mathcal{L} =
      \mathcal{L}_{\mathrm{Dice}}
      + \mathcal{L}_{\mathrm{CE}}
      + 0.5\,\mathcal{L}_{\mathrm{Bdry}}.
  \label{eq:loss}
\end{equation}
All three terms are weighted by the inverse voxel frequency
$w_c\!=\!|\Omega|\,/\,(N_{\mathrm{cls}}\cdot\sum_{\mathbf{r}}
y_{c,\mathbf{r}})$ to counteract foreground-background imbalance.
$\mathcal{L}_{\mathrm{Dice}}$ optimizes region overlap;
$\mathcal{L}_{\mathrm{CE}}$ provides per-voxel calibration.
$\mathcal{L}_{\mathrm{Bdry}}$ applies a masked binary
cross-entropy restricted to voxels within a morphologically dilated
($d\!=\!3$, $\ell_\infty$ norm) class-boundary region $\mathbf{M}$,
normalized by $|\mathbf{M}|$ to keep gradient magnitude consistent
across varying tumor sizes.

\section{Experimental Results}
\label{sec:exp-results}
We evaluate the proposed model, LightMedSeg, against state-of-the-art 3D medical image segmentation methods on two widely used volumetric benchmarks: BraTS (Brain Tumor Segmentation)~\cite{4b589b6824a64a2a91e8e3b26cc0bf9e} and ACDC (Automatic Cardiac Diagnosis Challenge)~\cite{acdc}. To ensure fair comparability with prior work, we  follow the training and preprocessing protocols of nnFormer~\cite{nnFormer}, including dataset splits, intensity normalization, cropping and evaluation metrics. No external data sources, pretraining, or auxiliary datasets are used.

We train LightMedSeg from scratch for 300 epochs with a batch size of 8 for BraTS and ACDC. During training, random 3D crops of size $128 \times 128 \times 128$ voxels are extracted from the volumes.  BraTS contains 411 training and 73 testing cases. Here, each modality is normalized via per volume $Z-score$. For ACDC intensities are clipped to the 0.5–99.5 percentile range and linearly rescaled to [0,1]. We follow the official dataset splits and apply standard 3D data augmentation, including random flipping, rotation, and Gaussian noise. The model parameters are optimized using the AdamW optimizer~\cite{loshchilov2019decoupledweightdecayregularization}, with an initial learning rate of $2\times 10^{-4}$, weight decay $1\times 10^{-5}$, and default betas (0.9, 0.999). A cosine annealing scheduler~\cite{cos} is applied with a minimum learning rate of $1\times 10^{-9}$ and $T_{\text{max}}=100$ epochs, following a short warmup. 

\begin{table*}[!t]
\centering
\small
\caption{Comparison of various methods on the BraTS dataset using average Dice score. Bold values represent the best performance, while underlined values indicate the second best results. Here, the number of parameters (denoted as 'params') is reported in millions. }

\begin{tabular}{l|c|c|c|c|c} 
\hline
\textbf{Methods} & \textbf{Param} & \textbf{Avg \%} & \textbf{Whole Tumor ↑} & \textbf{Enhancing Tumor ↑} & \textbf{Tumor Core ↑} \\
\hline
nnFormer\cite{nnFormer}             & 150.5  & \textbf{86.4}  & \textbf{91.3}  & \textbf{81.8}  & \textbf{86.0}  \\
Segformer3D\cite{10678245}          & 4.5    & 82.1  & 89.9  & 74.2  & 82.2  \\
UNETR\cite{UNETR}                & 92.49  & 71.1  & 78.9  & 58.5  & 76.1  \\
TransBTS\cite{TransBTS}            & --     & 69.6  & 77.9  & 57.4  & 73.5  \\
CoTr\cite{xie2021cotrefficientlybridgingcnn}                & 41.9   & 68.3  & 74.6  & 55.7  & 74.8  \\
CoTr w/o CNN Encoder\cite{xie2021cotrefficientlybridgingcnn} & --     & 64.4  & 71.2  & 52.3  & 69.8  \\
TransUNet\cite{chen2021transunettransformersmakestrong}            & 96.07  & 64.4  & 70.6  & 54.2  & 68.4  \\
SETR MLA\cite{DBLP:journals/corr/abs-2012-15840}             & 310.5  & 63.9  & 69.8  & 55.4  & 66.5  \\
SETR PUP\cite{DBLP:journals/corr/abs-2012-15840}            & 318.31 & 63.8  & 69.6  & 54.9  & 67.0  \\
SETR NUP\cite{DBLP:journals/corr/abs-2012-15840}             & 305.67 & 63.7  & 69.7  & 54.4  & 66.9  \\
\hline
\textbf{LightMedSeg(Ghost Conv)}            & \textbf{0.48}   & 83.4 & 89.8 & 77.0 & 83.5 \\
\textbf{LightMedSeg(Standard Conv)}            & 0.66   & \underline{84.8} & \underline{90.9} & \underline{80.3} & \underline{83.9} \\
\hline
\end{tabular}

\label{tab:brats_comparison}
\end{table*}

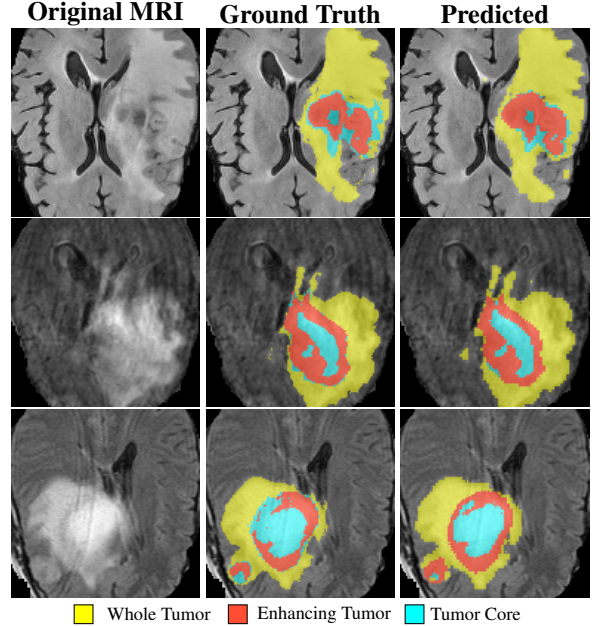
\begin{figure}[!hbt]
\centering

\begin{minipage}{0.3\linewidth}
  \centering
  \textbf{Original MRI}
\end{minipage}\hspace{0pt}
\begin{minipage}{0.3\linewidth}
  \centering
  \textbf{Ground Truth}
\end{minipage}\hspace{0pt}
\begin{minipage}{0.3\linewidth}
  \centering
  \textbf{Predicted}
\end{minipage}

\begin{minipage}{0.3\linewidth}
  \centering
  \includegraphics[angle=-90, width=2.5cm]{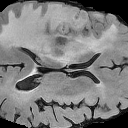}
\end{minipage}\hspace{0pt}
\begin{minipage}{0.3\linewidth}
  \centering
  \includegraphics[angle=-90,width=2.5cm]{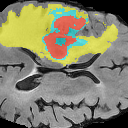}
\end{minipage}\hspace{0pt}
\begin{minipage}{0.3\linewidth}
  \centering
  \includegraphics[angle=-90,width=2.5cm]{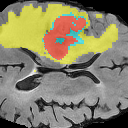}
\end{minipage}

\begin{minipage}{0.3\linewidth}
  \centering
  \includegraphics[angle=-90,width=2.5cm]{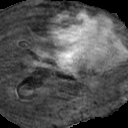}
\end{minipage}\hspace{0pt}
\begin{minipage}{0.3\linewidth}
  \centering
  \includegraphics[angle=-90,width=2.5cm]{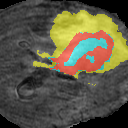}
\end{minipage}\hspace{0pt}
\begin{minipage}{0.3\linewidth}
  \centering
  \includegraphics[angle=-90,width=2.5cm]{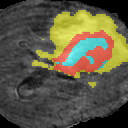}
\end{minipage}

\begin{minipage}{0.3\linewidth}
  \centering
  \includegraphics[angle=-90,width=2.5cm]{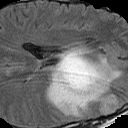}
\end{minipage}\hspace{0pt}
\begin{minipage}{0.3\linewidth}
  \centering
  \includegraphics[angle=-90,width=2.5cm]{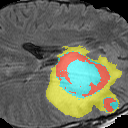}
\end{minipage}\hspace{0pt}
\begin{minipage}{0.3\linewidth}
  \centering
  \includegraphics[angle=-90,width=2.5cm]{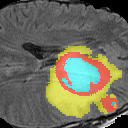}
\end{minipage}

\begin{tikzpicture}[baseline=(current bounding box.center)]
  \scriptsize
  \draw[fill={rgb,255:red,255; green,255; blue,0},draw=black] (0,0) rectangle (0.25,0.25);
  \node[anchor=west] at (0.35,0.125) {Whole Tumor};
  \draw[fill={rgb,255:red,255; green,77; blue,51},draw=black] (2.05,0) rectangle (2.3,0.25);
  \node[anchor=west] at (2.35,0.125) {Enhancing Tumor};
  \draw[fill={rgb,255:red,0; green,255; blue,255},draw=black] (4.4,0) rectangle (4.65,0.25);
  \node[anchor=west] at (4.65,0.125) {Tumor Core};
\end{tikzpicture}

\caption{Visual comparison of original MRI, ground truth, and predicted segmentation for four representative BraTS cases.}
\label{fig:brats_results}
\end{figure}

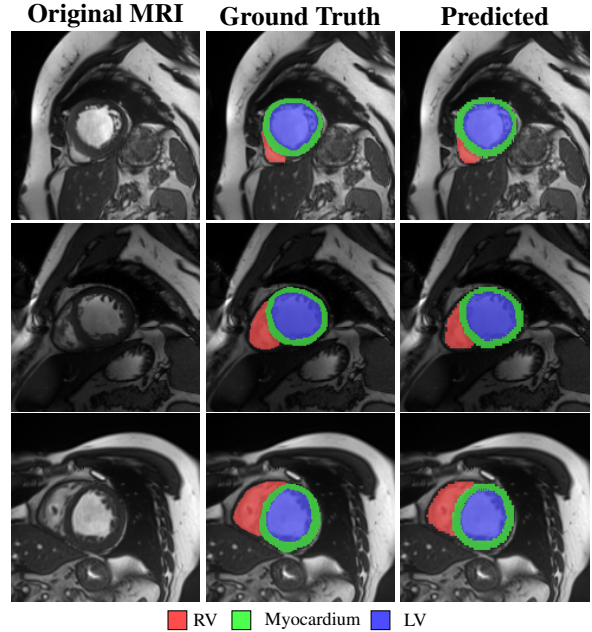
\begin{figure}[t]
\centering

\begin{minipage}{0.3\linewidth}
  \centering
  \textbf{Original MRI}
\end{minipage}\hspace{0pt}
\begin{minipage}{0.3\linewidth}
  \centering
  \textbf{Ground Truth}
\end{minipage}\hspace{0pt}
\begin{minipage}{0.3\linewidth}
  \centering
  \textbf{Predicted}
\end{minipage}

\vspace{0.05em}

\begin{minipage}{0.3\linewidth}
  \centering
  \includegraphics[width=2.5cm]{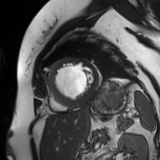}
\end{minipage}\hspace{0pt}
\begin{minipage}{0.3\linewidth}
  \centering
  \includegraphics[width=2.5cm]{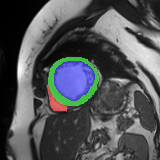}
\end{minipage}\hspace{0pt}
\begin{minipage}{0.3\linewidth}
  \centering
  \includegraphics[width=2.5cm]{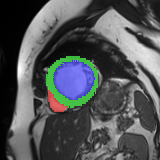}
\end{minipage}

\vspace{0.05em}

\begin{minipage}{0.3\linewidth}
  \centering
  \includegraphics[width=2.5cm]{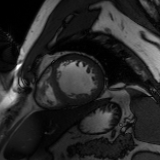}
\end{minipage}\hspace{0pt}
\begin{minipage}{0.3\linewidth}
  \centering
  \includegraphics[width=2.5cm]{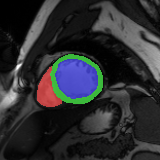}
\end{minipage}\hspace{0pt}
\begin{minipage}{0.3\linewidth}
  \centering
  \includegraphics[width=2.5cm]{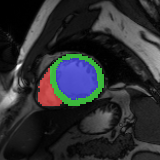}
\end{minipage}

\begin{minipage}{0.3\linewidth}
  \centering
  \includegraphics[width=2.5cm]{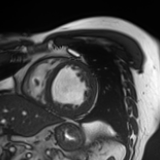}
\end{minipage}\hspace{0pt}
\begin{minipage}{0.3\linewidth}
  \centering
  \includegraphics[width=2.5cm]{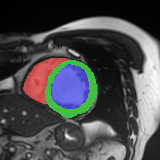}
\end{minipage}\hspace{0pt}
\begin{minipage}{0.3\linewidth}
  \centering
  \includegraphics[width=2.5cm]{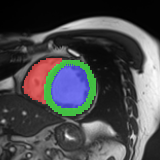}
\end{minipage}

\vspace{0.05em}

\begin{tikzpicture}[baseline=(current bounding box.center)]
  \scriptsize
  \draw[fill={rgb,255:red,255; green,77; blue,77},draw=black] (0.65,0) rectangle (0.90,0.25);
  \node[anchor=west] at (0.90,0.125) {RV};
  \draw[fill={rgb,255:red,77; green,255; blue,77},draw=black] (1.5,0) rectangle (1.75,0.25);
  \node[anchor=west] at (1.85,0.125) {Myocardium};
  \draw[fill={rgb,255:red,77; green,77; blue,255},draw=black] (3.35,0) rectangle (3.6,0.25);
  \node[anchor=west] at (3.7,0.125) {LV};
\end{tikzpicture}

\caption{Visual comparison of original MRI, ground truth, and predicted segmentation for four ACDC cases.}
\label{fig:acdc_results}
\end{figure}

\begin{table}[!hbt]
\centering
\small
\setlength{\tabcolsep}{4pt}
\caption{Comparison on ACDC dataset (avg.\ Dice score). Params in millions. \textbf{Bold}: best, \underline{underline}: second best.}
\begin{tabular}{l|c|c|c|c|c}
\hline
Method & Params & Avg\%↑ & RV & Myo & LV \\
\hline
nnFormer~\cite{nnFormer}         & 150.5 & \underline{92.06} & 90.94 & \underline{89.58} & 95.65 \\
Segformer3D~\cite{10678245}      & 4.5   & 90.96 & 88.50 & 88.86 & 95.53 \\
LeViT-UNet~\cite{LeViT-UNet}     & 52.17 & 90.32 & 89.55 & 87.64 & 93.76 \\
SwinUNet~\cite{swinUNet}         & --    & 90.00 & 88.55 & 85.62 & 95.83 \\
TransUNet~\cite{chen2021transunettransformersmakestrong} & 96.07 & 89.71 & 88.86 & 85.54 & 95.73 \\
UNETR~\cite{UNETR}               & 92.49 & 88.61 & 85.29 & 86.52 & 94.02 \\
R50-ViT~\cite{chen2021transunettransformersmakestrong}   & 86.00 & 87.57 & 86.07 & 81.88 & 94.75 \\
ViT-CUP~\cite{chen2021transunettransformersmakestrong}   & 86.00 & 81.45 & 81.46 & 70.71 & 92.18 \\
UNETR++~\cite{unetr++}           & 42.96 & \textbf{92.83} & \underline{91.89} & \textbf{90.61} & \textbf{96.00} \\
\hline
\textbf{LightMedSeg} & \textbf{0.48} & 91.24 & 91.77 & 86.24 & \underline{95.72} \\
\textbf{(Ghost Conv)} & & & & & \\[2pt]
\textbf{LightMedSeg} & 0.66 & 91.81 & \textbf{92.17} & 87.78 & 95.47 \\
\textbf{(Standard Conv)} & & & & & \\

\hline
\end{tabular}
\label{table:acdc}
\end{table}

The performance is reported using Dice score as the primary metric for the relative comparisons of models. Tables~\ref{tab:brats_comparison} and~\ref{table:acdc} present the results on BraTS and ACDC, respectively. On BraTS, LightMedSeg achieves an average Dice score of 83.4\% using only 0.48M parameters, compared to nnFormer, which achieves 86.4\% with 150M parameters over 50$\times$ larger. On ACDC, LightMedSeg achieves 91.24\%, which is close to the best performer, UNETR++, with 92.83\%.

Figures~\ref{fig:brats_results} and~\ref{fig:acdc_results} visualize representative predictions. On BraTS, our model accurately captures small enhancing tumor regions and irregular boundaries, while on ACDC it produces precise LV and myocardium masks with sharp boundaries. These results indicate strong clinical potential under both limited compute and limited data regimes. 

\begin{table}[t]
\centering
\caption{Comparison of LightMedSeg with state-of-the-art methods
in terms of model size, computational complexity, and peak memory.
Measured on a BraTS-sized input
($1\!\times\!4\!\times\!128^3$); {--} indicates unavailable data.}
\begin{tabular}{@{} l r r r @{}}
\toprule
Architecture & \makecell{Params \\ (M)} & \makecell{FLOPs \\ (G)} &
               \makecell{Mem \\ (MB)} \\
\midrule
nnFormer~\cite{nnFormer}       & 150.5 & 213.4 & 12902 \\
UNETR~\cite{UNETR}             & 92.49 & 75.76 & 3379  \\
SwinUNETR~\cite{10.1007/978-3-031-08999-2_22} & 62.83 & 384.2 & 20173 \\
UNETR++~\cite{unetr++}         & 42.96& 70.1 &
                                 \underline{2458}  \\
SegFormer3D~\cite{10678245}    & \underline{4.51}  & \underline{17.5}  & {--}  \\
\addlinespace
\textbf{LightMedSeg}           & \textbf{0.48} & \textbf{14.64} &
                                 \textbf{1174}  \\
\bottomrule
\end{tabular}
\label{table2}
\end{table}
As shown in Table~\ref{table2}, LightMedSeg achieves the lowest
parameter count 0.48M and FLOPs 14.64~GFLOPs among all
compared architectures. It is $9\times$ more compact than the next
most efficient baseline SegFormer3D and over $88\times$ smaller
than UNETR++. Owing to the absence of self-attention and extensive use of
depthwise and ghost convolutions, LightMedSeg processes a
$128^3$ volume in approximately 13.7~ms on a single NVIDIA
RTX 5080 GPU and 505.4 ms on Ryzen 9 9950x CPU, demonstrating suitability for real-time clinical
deployment.

\begin{table}[t]
\caption{Component ablation on BraTS (Avg.\ Dice). $K$ denotes
the number of anatomical anchors.}

\centering
\small
\setlength{\tabcolsep}{6pt}
\begin{tabular}{lc}
\toprule
Configuration       & Avg Dice \\
\midrule
Full Model          & \textbf{83.43} \\
w/o LSPM            & 80.50 \textcolor{gray}{(\(-2.93\))} \\
w/o Anchors         & 81.93 \textcolor{gray}{(\(-1.50\))} \\
w/o Skip Router     & 82.27 \textcolor{gray}{(\(-1.16\))} \\
w/o GhostConv3D     & 84.81 \textcolor{gray}{(\(+1.38\))} \\
\midrule
$K{=}8$ (default)   & \textbf{83.43} (0.48M) \\
$K{=}16$            & 83.64 (0.49M) \\
$K{=}32$            & 84.16 (0.52M) \\
\bottomrule
\end{tabular}
\label{tab:ablation}
\end{table}

Table~\ref{tab:ablation} shows the ablation results. These results highlight that LSPM contributes the most to segmentation accuracy, indicating that explicitly modeling structural complexity is crucial for lightweight architectures. Removing the LSPM causes the largest drop ($-2.93$ Dice), confirming structural priors and texture-aware routing as the most critical choices; disabling the Global Anchor Detector and the learned skip router
cost $-1.50$ and $-1.16$ Dice respectively. Removing GhostConv3D
recovers $+1.38$ Dice at the cost of more parameters. $k$ is the number of
anatomical anchors, we default to $K\!=\!8$ giving us 83.43 Dice over $K\!=\!32$ with 84.16, to best suit our efficiency target.

\section{Conclusion}
\label{sec:con}

We presented LightMedSeg, a compact 3D medical image
segmentation network that achieves accuracy competitive with models
orders of magnitude larger. By combining anchor conditioned FiLM
modulation, local structural priors, texture-aware routing, and
learned multi-scale skip fusion, the model delivers strong
performance on BraTS and ACDC while remaining deployable
under strict memory and compute constraints. Future work will explore
semi-supervised and multi-institutional settings, as well as
extension to additional imaging modalities.

\end{document}